\documentclass[10pt,twocolumn,letterpaper]{article}

\usepackage{cvpr}
\usepackage{times}
\usepackage{epsfig}
\usepackage{graphicx}
\usepackage{amsmath}
\usepackage{amssymb}
\usepackage{multirow}
\usepackage{diagbox}
\usepackage{animate}
\usepackage{mathrsfs}
\usepackage[table,xcdraw]{xcolor}
\usepackage{subfig}
\usepackage{slashbox}
\usepackage{tabularx}
\usepackage{booktabs}
\usepackage{enumitem}

\usepackage[pagebackref=true,breaklinks=true,letterpaper=true,colorlinks,bookmarks=false]{hyperref}

\cvprfinalcopy 

\def\cvprPaperID{3106} 

\ifcvprfinal\fi
\begin{document}
\title{\emph{What} and \emph{How Well} You Performed? A Multitask Learning Approach to \\Action Quality Assessment}
\author{Paritosh Parmar \hspace{2cm} Brendan Tran Morris \\
University of Nevada, Las Vegas\\
{\tt\small parmap1@unlv.nevada.edu, brendan.morris@unlv.edu}
}
\maketitle
\begin{abstract}
Can performance on the task of action quality assessment (\textsc{\textit{AQA}}) be improved by exploiting a description of the action and its quality? Current \textsc{\textit{AQA}} and skills assessment approaches propose to learn features that serve only one task - estimating the final score. In this paper, we propose to learn spatio-temporal features that explain three related tasks - fine-grained action recognition, commentary generation, and estimating the \textsc{\textit{AQA}} score.
A new multitask-\textsc{\textit{AQA}} dataset, the largest to date, comprising of 1412 diving samples was collected to evaluate our approach (\url{https://github.com/ParitoshParmar/MTL-AQA}). We show that our \textsc{\textit{MTL}} approach outperforms \textsc{\textit{STL}} approach using two different kinds of architectures: \textsc{\textit{C3D-AVG}} and \textsc{\textit{MSCADC}}.  The \textsc{\textit{C3D-AVG-MTL}} approach achieves the new state-of-the-art performance with a rank correlation of 90.44\%.  Detailed experiments were performed to show that \textsc{\textit{MTL}} offers better generalization than \textsc{\textit{STL}}, and representations from action recognition models are not sufficient for the \textsc{\textit{AQA}} task and instead should be learned.
\end{abstract}
\section{Introduction}
\label{sec:intro}
What score should an athlete receive on her dive/gymvault/skating/etc? Which med student has the highest surgical skill level?  How well can he paint or draw?  How is a patient progressing in their physical rehabilitation program?  Answering these questions involves the quantification of the quality of the action -- determining \textit{how well} the action was carried out, also known as action quality assessment (AQA).
\begin{figure}[h]
\centering
\captionsetup[subfigure]{labelformat=empty} 
\subfloat[]{}
\animategraphics[autoplay,loop,poster=17,height=0.15\textwidth]{3}{videos/d4/image_}{0001}{0030} 
\includegraphics[width=\linewidth]{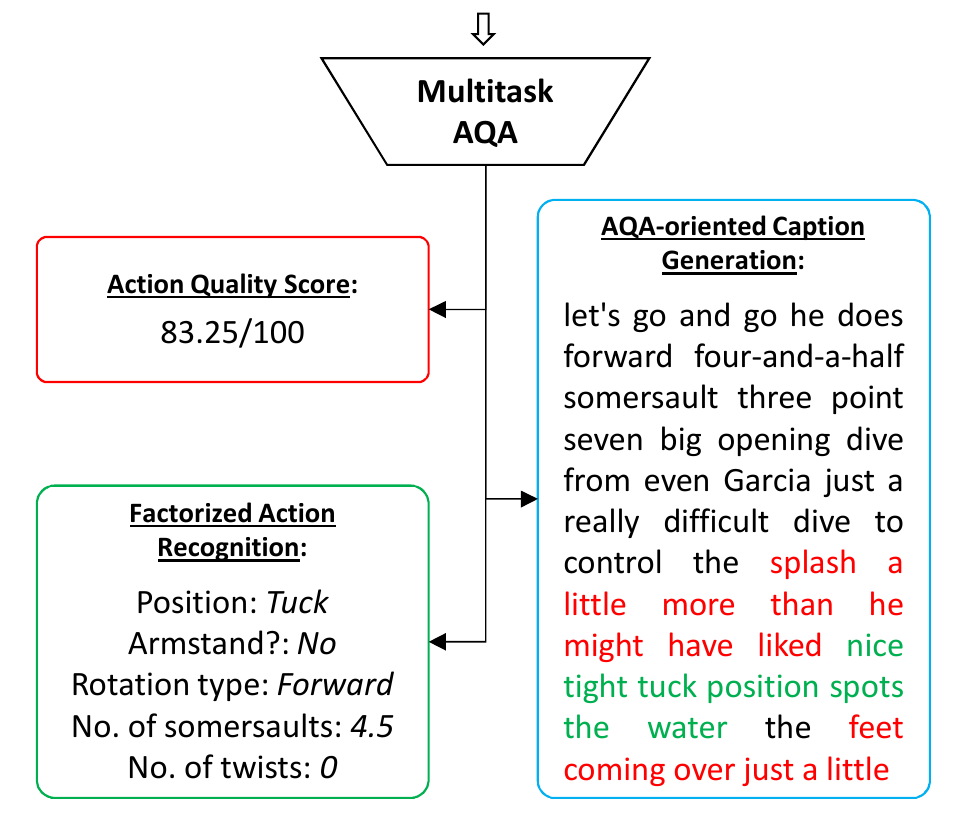}
\caption{\textbf{Multitask AQA concept}. 
Recognizing an action instance in detail and verbally describing its good and bad points can be helpful in the process of quantifying the quality of that action instance. We propose to learn a model that delineates an action besides measuring its quality. \textit{To see the videos play, please download the manuscript and view in an Adobe Reader.}}
\label{fig:concept}
\end{figure}
Existing AQA \cite{pirsia, parmar17, s3d, yongjun, venkat} and skills assessment \cite{doughty17, fawaz, zia18, zia18_1, zia16} approaches use a single label, known as a final score or skill-level, to train the system using some kind of regression or ranking loss function.  However, the performance of these systems is limited and it seems that a single score is not sufficient to characterize a complicated action. In AQA, the final score is dependent on what was done (this determines the difficulty level) and how was that done (this determines the quality of execution). We pose the following question: \textit{can learning to describe and commentate on the action instances help improve the performance on the AQA task?}   

We hypothesize that by forcing the network to learn to do so will help better characterize the action, and hence aid in AQA. So, rather than using just a single encompassing quality label to train the network, we introduce a multitask learning (MTL) approach (Fig. \ref{fig:concept}) to assess the quality of an action. Specifically, we propose to utilize 3D CNN's to learn spatio-temporal representations of salient motion and appearance; optimize those using loss functions which account for i) the action quality score, ii) factorized (detailed) action classification, and iii) generate a verbal commentary of performance; and are trained end-to-end. Note that the architectures are multitask and not multi-modal since the input does not use captions or action classification to produce the AQA score.  Besides straight forward utility for AQA and action classification, automatic commentary or sports narrative generation has been viewed valuable and greatly applicable in a recent work by Yu \etal \cite{narrative}.

For AQA tasks, domain experts can provide detailed analysis of performance.  In the professional sports setting, ground truth annotations for detailed action classification and commentary by former athletes are readily available in broadcast footage facilitating extraction of labels and descriptive captions.  As such, to evaluate our approach, we introduce the first multitask AQA dataset with 1412 samples of diving which is also the largest AQA dataset to date.

Experimental evaluation show that performance of both the architectures improved as more tasks were added and the C3D-AVG-MTL variant outperforms all existing AQA approaches in literature.  MTL was shown to outperform STL across various training set sizes.  Further experiments explore the AQA-orientedness of the feature representations learned by our networks and find they outperform action-recognition representations on unseen actions indicating that better generalized concepts of quality were learned.  

Contributions: primary novelty of this works lies in the problem formulation -- to learn spatio-temporal representations by optimizing networks end-to-end jointly for fine-grained action description and AQA scoring. Task selection is intuitive. No previous work has done this; not just for AQA, but even action recognition and captioning tasks. We release a novel MTL-AQA dataset which is the largest AQA dataset so far, much more diverse, challenging, and richly annotated with factorized fine-grained action class and AQA-oriented captions; can help researchers in the field to examine new ideas for AQA and auxiliary tasks. We show that our MTL approach works across different architectures. Our approach is applicable to a wide range of problems. Proposed models are simple, yet intuitive, and effective in carrying out central idea. Our C3D-AVG-MTL surpasses all the existing approaches.
\begin{table*}[t]
\small
\centering
\begin{tabular}{l|ccccccc}
\toprule
\textbf{Dataset}                                                                    & \textbf{Events}                                                    & \textbf{Height}                                                        & \textbf{Genders}                                       & \textbf{\# Samples} & \textbf{Events} & \textbf{\begin{tabular}[c]{@{}c@{}}View Variation/\\ Background\end{tabular}} & \textbf{Labels}                                                                                  \\ \midrule
\textbf{MIT Dive \cite{pirsia}} & Individual                                                         & 10m Platform                                                           & Male                                                   & 159                 & 1          & No/Same                                                                       & AQA score                                                                         \\ \midrule
\textbf{UNLV Dive \cite{parmar17}}     & Individual                                                         & 10m Platform                                                           & Male                                                   & 370                 & 1 & No/Same                                                                       & AQA score                                                                         \\ \midrule
\textbf{Ours MTL-AQA}                                                                       & \begin{tabular}[c]{@{}c@{}}Individual, \\ Synchronous\end{tabular} & \begin{tabular}[c]{@{}c@{}}3m Springboard,\\ 10m Platform\end{tabular} & \begin{tabular}[c]{@{}c@{}}Male,\\ Female\end{tabular} & 1412                & 16        & Yes/Different                                                                 & \begin{tabular}[c]{@{}c@{}}AQA score,\\ Action class,\\ Commentary\end{tabular} \\ \bottomrule
\end{tabular}
\caption{\textbf{Details of our newly introduced dataset}, and its comparison with the existing AQA datasets.}
\label{tab:dataset}
\end{table*}
\section{Related Work}
\label{sec:related}
\paragraph{AQA:}
Pirsiavash \etal \cite{pirsia} proposed the use of DFT/DCT of body pose as features for a support vector regressor (SVR) to map to a final action quality score.  They introduced an action quality dataset containing two actions: Diving and Figure Skating.  However, since their method relied solely on pose features, it neglected important visual quality cues, like splash in the case of Diving. Since accurate pose is especially difficult in sports scenarios where athletes undergo extremely convoluted poses, Venkataraman \etal \cite{venkat} better encoded using the approximate entropy of the poses to improve the results. 

More recently, spatio-temporal features from 3D convolutional neural networks (C3D) \cite{c3d} proved to be very successful on a related task of action recognition since they captured appearance and salient motion.  Seeing this as a desirable property that would help to take into account visual cues, Parmar and Morris \cite{parmar17} proposed using C3D features for AQA.  They proposed three frameworks, C3D-SVR, C3D-LSTM, and C3D-LSTM-SVR, which differed in their feature aggregation and regression scheme.  All the frameworks worked better than previous models proving the efficacy of C3D features for AQA. Xiang \etal \cite{s3d} proposed breaking video clips into action specific segments and fusing segment-averaged features instead of over full videos.  By adding finer segment labels to data samples performance was improved.  Li \etal \cite{yongjun} divide a sample into 9 clips and use 9 different C3D networks dedicated to different stags of Diving.  Features are concatenated and further processed through \texttt{conv} and \texttt{fc} layers to produce a final AQA score using a ranking loss along with the more typical L2 loss. Xu \etal \cite{xu18} tackle AQA for longer action sequences using self-attentive and multiscale convolutional skip LSTM.
\paragraph{Skills assessment:}
Zia \etal \cite{zia16} extract spatio-temporal interest points (STIP's) in the frequency domain to classify a sample into novice, intermediate or expert skills level. Instead of using handcrafted STIP's Doughty \etal \cite{doughty17} learn and use convolutional features with ranking loss as their objective function to evaluate surgical, drawing, chopstick use and dough rolling skills. In their subsequent work \cite{doughty18}, they use temporal attention. Li \etal \cite{li19manipulation}, make use of spatial attention in the assessment of hand manipulation skills. Bertasius \etal \cite{baller} focus on measuring basketball skills but rely only on assessment of a single basketball coach making their dataset subjective to a particular evaluator. 

All of the existing AQA and skills assessment frameworks are single task models and only give the final AQA score. Our proposed framework is a multitask model to recognize the action, measures its quality and also generates captions (or commentary).
\paragraph{Multi-modal approaches and captioning:}
Images and videos (especially sports) are often accompanied by a caption or commentary which can themselves serve as labels yet to be exploited for AQA or skill assessment. Quattoni \etal \cite{quattoni} use large quantities of unlabeled images, with associated captions, to learn image representations. They found that this sort of pre-training with extra information could speed up the learning on a target task.  Rather than using captions as groundtruth labels, Sonal \etal \cite{sonalecml} treated captions as a ``view'' and use them along with images to learn a classifier using co-training. They again used commentary as a ``view'' for action recognition with success.  To train an activity classifier in an automated fashion, without the requirement of any manual labeling, Sonal and Mooney \cite{sonal09} make use of broadcast closed captions and used the system for video retrieval. There are a few works which focus on captioning in sports settings. Yu \etal \cite{narrative} address the task of generating fine-grained video descriptions for basketball and evaluate performance using their novel metric. Commentary generation in cricket has been addressed in \cite{commbox, cricketcomm}, while Sukhwani addressed the problem of describing tennis videos in \cite{sukhwani}. While these works focus on captioning or improving captioning, we integrate a captioning task with an AQA task to provide stronger supervision as commentary is a verbal description of AQA.
\section{Multitask AQA Dataset}
\label{sec:dataset}
\begin{table}[]
\small
\centering
\begin{tabular}{ccccc}
\toprule
\textbf{Position}                                            & \textbf{Armstand}                                 & \textbf{\begin{tabular}[c]{@{}c@{}}Rotation type\end{tabular}}                 & \textbf{\# SS} & \textbf{\# TW} \\ \midrule
\begin{tabular}[c]{@{}c@{}}Free\\ Tuck\\ Pike\end{tabular} & \begin{tabular}[c]{@{}c@{}}No\\ Yes\end{tabular} & \begin{tabular}[c]{@{}c@{}}Inward\\ Reverse\\ Backward\\ Forward\end{tabular} & 0 to 4.5          & 0 to 3.5          \\ \bottomrule
\end{tabular}
\caption{\textbf{Classification of dives}. Each combination of the presented sub-fields produces a different kind of maneuver.}
\label{tab:dive_types}
\end{table}
In order to facilitate research in the area of AQA, we release a new dataset. This is the first of a kind multitask AQA dataset. With 1412 samples, it is the largest AQA dataset to date. This particular dataset focuses only on Diving as it has seen the most usage recently.  Data was compiled from 16 different events unlike the single main event (2012 Olympics Men's 10m Platform Diving competition) used for previous datasets \cite{pirsia, parmar17} to provide significantly more variation.  Diving samples in the new dataset were collected from various International competitions and include the 10m Platform as well as 3m Springboard, include both male and female athletes, individual or pairs of synchronized divers, and different views.  A comparison of our new dataset with existing Diving AQA datasets is provided in Table \ref{tab:dataset}.  

Since data was collected from televised international events, before the athletes perform their routines, information regarding their routine is displayed.  This information includes the difficulty of the dive and a description of the dive.  The AQA score is extracted from the judges' scores after the dive completion.  The dataset uses the same dive classification strategy as Nibali \etal \cite{nibali}, where instead of using dive number (equivalent to an action class in action recognition) directly, we factorize a dive into its components such as the position of the dive, the number of somersaults (SS), and number of twists (TW).  Full details for the dive classification is in Table \ref{tab:dive_types}.  

Further, during and after a diving routine, television analysts provide commentary.  These analysts are often retired athletes and have deep understanding of the sport.  This verbal account of the athlete's performance is recorded for the third type of action label.  The commentary was considered an important indicator for performance since it was the only way to ``watch'' an event before telecast was available.  Commentators say what the athlete performed, what was correct with the athlete's performance, and where and how athletes made mistakes. This provides deeper insight into the athlete's performance and can help an average person better understand the sport. We used Google's Speech-To-Text API to convert commentary audio to text. 
\section{Multitask Approach to AQA}
\label{sec:approach}
MTL is a machine learning paradigm in which a single model caters to more than a single task. An example is to recognize road signs, roads, and vehicles together while an STL approach would require separate models for each object type.  MTL tasks are generally chosen such that they are related to one another and their networks have a common body that branches into task-specific heads.  The total network loss is the sum of individual task losses.  When optimized end-to-end, the network is able to learn richer representation in the common body section since it must be able to serve/explain all tasks.  With the use of related auxiliary tasks, which are complementary to the main task, the richer representation tends to help improve performance on the main task.  

In general, not just for diving, action quality is a function of \emph{what} action was carried out and \textit{how well} that action was executed. This makes the choice of auxiliary tasks natural: detailed action recognition is the answer to the \textit{`what'} part and commentary, being a verbal description containing good and bad points about action execution, is an answer to the \textit{`how well'} part. AQA can be thought of as finding a function that maps input video to the AQA scores. Caruana in \cite{caruana97} views supervision signals from auxiliary tasks as an inductive bias (assumptions). Inductive bias can be thought of as constraints that restrict the hypothesis/search space when finding the AQA function. Through inductive biases, MTL provides improved generalization as compared STL \cite{caruana97}.
 
In this work, the main task is to assess the action quality (AQA score) and the auxiliary tasks are to recognize the action (dive type classification) and to generate descriptive captions/commentary.  Action recognition in turn consists of five fine-grained dive sub-recognition tasks: recognizing position and rotation type, detecting armstand, and counting somersaults and twists. 

First, let us formalize the settings and objective functions. AQA is a regression problem where, generally, the Euclidean distance between the predicted quality score and the ground truth is used as the objective function to be minimized \cite{parmar17, s3d, yongjun}. Initial experimentation found that using L1 distance in addition to L2 yielded better results on the AQA task
\begin{equation}
\mathcal{L}_{AQA} = -\frac{1}{N} \sum_{i=1}^{N} (x_{i}-y_{i})^2 + |x_{i}-y_{i}| \label{eq:1}
\end{equation}
where $x_i$ is the predicted score and $y_i$ is the ground truth score for each of the $N$ samples.  For action recognition, we use cross-entropy loss between the predicted labels and ground truth label
\begin{equation}
\mathcal{L}_{Cls} =  -\frac{1}{N} \sum_{i=1}^{N} \sum_{sa} \sum_{j=1}^{k_{sa}} y^{sa}_{i,j}log(x^{sa}_{i,j}) \label{eq:2}
\end{equation}
where $k_{sa}$ is the number of categories in sub-action class $sa$ (as in Table \ref{tab:dive_types}).  Negative log likelihood is used as the loss function for the captioning task 
\begin{equation}
\mathcal{L}_{Cap} = -\frac{1}{N} \sum_{i=1}^{N} \sum_{sl} ln({x^{cap}}_{y^{cap}}) \label{eq:3}
\end{equation}
with $sl$ is the sentence length.  The overall objective function to be minimized is the summation of all the losses
\begin{equation}
\mathcal{L}_{MTL} = \alpha\mathcal{L}_{AQA} + \beta\mathcal{L}_{AR} + \gamma\mathcal{L}_{Cap}. \label{eq:4}
\end{equation} 
where $\alpha, \beta, \gamma$ are loss the weights. Now, we will introduce two different architectures for MTL-AQA. 
\paragraph{MTL-AQA architectures} Unlike action recognition that may be accomplished by looking at as little evidence as just a single frame \cite{karpathy14}, for AQA the complete action sequence needs to be considered because the athlete can make or lose points at any point during the whole sequence. 

While spatio-temporal representations learnt using 3D CNN's capture appearance and salient motion patterns \cite{c3d}, which makes them one of the best candidates for action recognition \cite{c3d, hara3d} and also for AQA \cite{parmar17, s3d, yongjun}, 3D CNN's require large memories which limits their application to small clips. We tackle this bottleneck in two ways:  
\begin{enumerate}[noitemsep,nolistsep]
\item divide the video (96 frames) into small clips (16 frames), and then aggregate clip-level representations to obtain video-level description (Sec. \ref{approach:avg_aggre})
\item downsample the video into a small clip (Sec. \ref{approach:mscadc})
\end{enumerate}

Networks designed for multitask learning generally two segments: \textbf{common network backbone} and \textbf{task-specific heads}. Common network backbone learns shared representations, which are then further processed through task-specific heads to obtain more task-oriented features and outputs.
\subsection{Averaging as aggregation (\textbf{\textsc{C3D-AVG}})}
\label{approach:avg_aggre} 
\begin{figure*}[t]
\centering
\includegraphics[width=\textwidth]{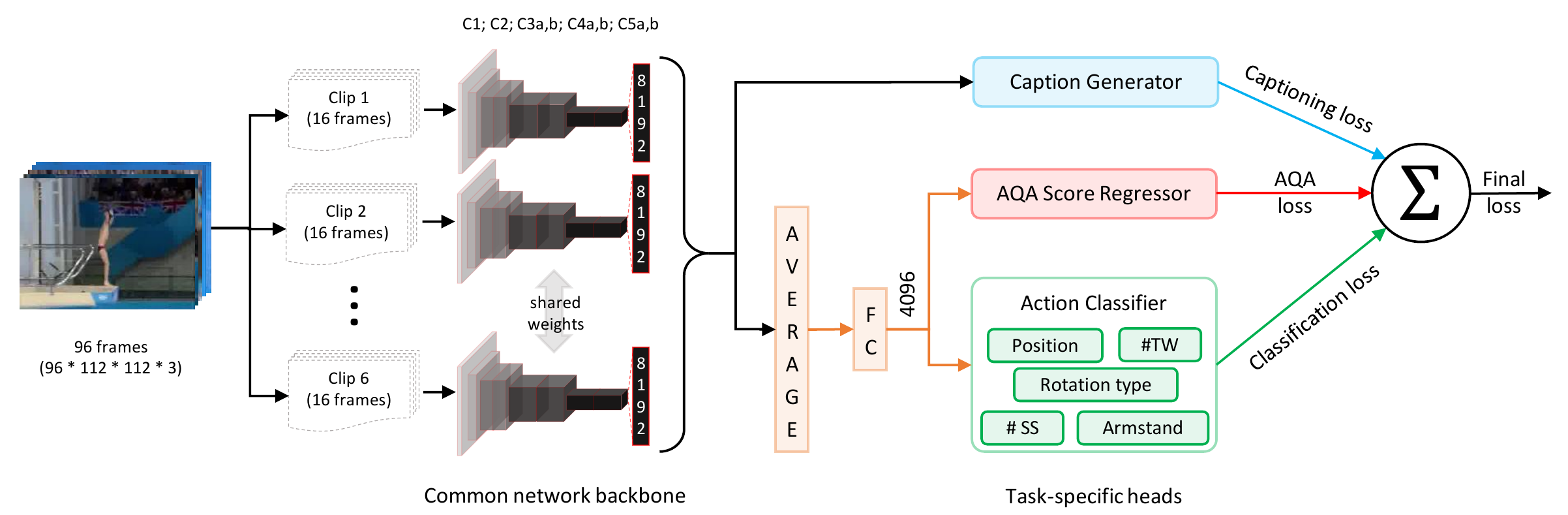}
\caption{\textbf{\textsc{C3D-AVG-MTL} network}.}
\label{fig_c3d_lstm_avg}
\end{figure*}   
The first network we present is C3D-AVG (Fig. \ref{fig_c3d_lstm_avg}). \\
\textbf{Network backbone:} Backbone consists of C3D network \cite{c3d} up to the fifth pooling layer. \\
\textbf{Aggregation scheme:} An athlete gathering (or losing) points throughout the action can be seen as an addition operation. Combining this perspective with a good rule of thumb that when good representations are learned, linear operations on them become meaningful, we propose to enforce a linear combination of representations to be meaningful, in order to learn good representations. Specifically, we propose to use \textit{averaging} as the linear combination.
The network is optimized end-to-end for all three tasks.

C3D-AVG network up to \texttt{Average} layer can be considered as an encoder, which encodes input video-clips into representations that when averaged (in feature space) would correspond to the total AQA points gathered by the athlete. Subsequent layers can be thought of decoders for individual tasks.\\
\textbf{Task-specific heads:} For action-recognition and AQA tasks, clip-level \texttt{pool-5} features are averaged element-wise to yield a video-level representation. Since captioning is a sequence-to-sequence task, the individual clip-level features are input to the captioning branch before averaging (individual clip-level features worked better in practice than averaged clip-level features for captioning). 
\subsection{Multiscale Context Aggregation with Dilated Convolutions (\textbf{\textsc{MSCADC}})} 
\label{approach:mscadc}
Multiscale context aggregation with dilated convolutions (MSCADC) \cite{contextnet} has been shown to improve the classification of dives in the work of Nibali \etal \cite{nibali}. Given its strong performance on an auxiliary task MSCADC was selected for MTL.  Our MTL variant network has a backbone and multiple heads as illustrated in Table \ref{tab:archi_dilated}. \\ \textbf{Network backbone: } The MSCADC network is based on C3D network \cite{c3d} and incorporates improvements like using Batch Normalization \cite{bn} to provide better regularization which is needed in AQA where data is quite limited. Additionally, pooling is removed from the last two convolutional groups of C3D and instead a dilation rate of 2 is used.  This backbone structure is shared among all the MTL tasks.\\
\textbf{Task-specific heads: } We use separate heads, one for each task. Heads consist of a context net followed by a few additional layers. The context net is where the feature maps are aggregated at multiple scales. 

Dilated convolutions and multi-scale aggregation have shown improvements in the tasks involving dense predictions \cite{contextnet}. We believe that removing pooling layers and using dilated convolutions better maintains the structure of the diving athlete without losing resolution. This helps in better assessment of  the athlete's pose which is critical for AQA. For example, pose can identify when legs are aligned or split which is useful not only for diving but also other sports such as gymnastic vault, figure skating, skiing, snowboarding, etc.

Unlike the C3D-AVG network, we downsample the complete action into a short sequence of only 16 frames (something like key action snapshots) as done by Nibali \etal \cite{nibali}. This reduces our 96-frames videos into key action snapshots which helps in processing the complete action sequence in a single pass.  Processing an action sequence using this network can be thought of as distilling information from the input frames and putting it into feature maps, with different feature maps containing different kinds of pose information. A natural benefit of downsampling the sequence is that there is a significant reduction in the the number of network parameters and memory which can be used instead to increase spatial resolution.
\begin{table}[]
\small
\centering
\begin{tabular}{c|c|c}
\cline{2-2}
\textit{\textbf{}}                                                                                                         & \textit{\textbf{(Common network body)}}                                                                                               &                                                                                                                            \\ \cline{2-2}
\cellcolor[HTML]{FFFFFF}                                                                                                   & \cellcolor[HTML]{ECF4FF}C3(32); BN                                                                                                    &                                                                                                                            \\ 
\cellcolor[HTML]{FFFFFF}                                                                                                   & \cellcolor[HTML]{ECF4FF}MP(1,2,2)                                                                                                     &                                                                                                                            \\ 
\cellcolor[HTML]{FFFFFF}                                                                                                   & \cellcolor[HTML]{ECF4FF}C3(64); BN                                                                                                    &                                                                                                                            \\ 
\cellcolor[HTML]{FFFFFF}                                                                                                   & \cellcolor[HTML]{ECF4FF}MP(2,2,2)                                                                                                     &                                                                                                                            \\ 
\cellcolor[HTML]{FFFFFF}                                                                                                   & \cellcolor[HTML]{ECF4FF}\{C3(128); BN\} x2                                                                                            &                                                                                                                            \\ 
\cellcolor[HTML]{FFFFFF}                                                                                                   & \cellcolor[HTML]{ECF4FF}MP(2,2,2)                                                                                                     &                                                                                                                            \\ 
\cellcolor[HTML]{FFFFFF}                                                                                                   & \cellcolor[HTML]{ECF4FF}\{C3(256); BN\} x2                                                                                            &                                                                                                                            \\ 
\cellcolor[HTML]{FFFFFF}                                                                                                   & \cellcolor[HTML]{ECF4FF}\{C3(d=2,256); BN\} x2                                                                                        &                                                                                                                            \\ 
\cellcolor[HTML]{FFFFFF}                                                                                                   & \cellcolor[HTML]{ECF4FF}Dropout(0.5)                                                                                                  &                                                                                                                            \\ \cline{2-2}
\textit{\textbf{}}                                                                                                         & \textit{\textbf{(Task-specific heads)}}                                                                                               &                                                                                                                            \\ \hline
\multicolumn{1}{|c|}{\cellcolor[HTML]{FBE8E7}\textit{\textbf{\begin{tabular}[c]{@{}c@{}}(AQA Score\\ Head)\end{tabular}}}} & \cellcolor[HTML]{EDFFED}\textit{\textbf{\begin{tabular}[c]{@{}c@{}}(Action \\ recognition Head)\end{tabular}}}                        & \multicolumn{1}{c|}{\cellcolor[HTML]{D4F5FF}\textit{\textbf{\begin{tabular}[c]{@{}c@{}}(Captioning\\ Head)\end{tabular}}}} \\ \hline
\multicolumn{1}{|c|}{\cellcolor[HTML]{FBE8E7}C1(12)}                                                                       & \cellcolor[HTML]{EDFFED}C1(12)                                                                                                        & \multicolumn{1}{c|}{\cellcolor[HTML]{D4F5FF}C1(12)}                                                                        \\ 
\multicolumn{1}{|c|}{\cellcolor[HTML]{FBE8E7}\{Cntxt net\}}                                                                & \cellcolor[HTML]{EDFFED}\{Cntxt net\}                                                                                                 & \multicolumn{1}{c|}{\cellcolor[HTML]{D4F5FF}\{Cntxt net\}}                                                                 \\ 
\multicolumn{1}{|c|}{\cellcolor[HTML]{FBE8E7}MP(2,2,2)}                                                                    & \cellcolor[HTML]{EDFFED}MP(2,2,2)                                                                                                     & \multicolumn{1}{c|}{\cellcolor[HTML]{D4F5FF}MP(2,2,2)}                                                                     \\ 
\multicolumn{1}{|c|}{\cellcolor[HTML]{FBE8E7}C3(12); BN}                                                                   & \cellcolor[HTML]{EDFFED}C3(12); BN                                                                                                    & \multicolumn{1}{c|}{\cellcolor[HTML]{D4F5FF}C3(12); BN}                                                                    \\ \hline
\multicolumn{1}{|c|}{\cellcolor[HTML]{FBE8E7}C3(1)}                                                                        & \cellcolor[HTML]{EDFFED}                                                                                                              & \multicolumn{1}{c|}{\cellcolor[HTML]{D4F5FF}Enc. GRU}                                                                      \\ 
\multicolumn{1}{|c|}{\cellcolor[HTML]{FBE8E7}AP(2,11,11)}                                                                  & \multirow{-2}{*}{\cellcolor[HTML]{EDFFED}\textit{\textbf{\begin{tabular}[c]{@{}c@{}}(Action \\ recognition sub-heads)\end{tabular}}}} & \multicolumn{1}{c|}{\cellcolor[HTML]{D4F5FF}Dec. GRU}                                                                      \\ \hline
\end{tabular}
\caption{\textbf{\textsc{MSCADC-MTL} architecture}. C3(d,ch): 3D convolutions, ch-no. of channels, d-dilation rate. C1: 1 $\times$ 1 $\times$ 1 convolutions. BN: batch normalization. MP(kr): max pooling operation, kr-kernel size. Cntxt net: context net for multi-scale context aggregation. AP: average pooling across (2 $\times$ 11 $\times$ 11) volume.}
\label{tab:archi_dilated}
\end{table}

\section{Experiments}
\label{sec:experiments}
\paragraph{Implementation:} PyTorch \cite{pytorch} is used to implement all the networks; common network backbones were pretrained on the UCF101 \cite{ucf101} action recognition dataset.  The captioning module utilized a GRU \cite{gru} cell and a dropout rate of 0.2 in the encoder and decoder. Maximum caption length is set to 100 words. Full vocabulary size is 5779. The parameters $\alpha$, $\beta$, and $\gamma$ in Eq. \ref{eq:4} are set to 1, 1, and 0.01. All networks used the Adam optimizer \cite{adam} and were trained for 100 epochs with initial learning rate of 1e-4.  Data augmentation is performed through center cropping with temporal augmentation and random horizontal flipping.  The center crop was found to reliably capture both the athlete and other prominent visual cues such as splash.  Batch-size was set to three samples.  Additional architecture-specific implementation details are as follows:\\
\textbf{C3D-AVG:} The model is trained end-to-end with a $112\times112$ center crop from the $171\times128$ pixel input video. Each dive sample was temporally normalized to a length of 96 frames.\\
\textbf{MSCADC:} Since this architecture does not contain fully-connected layers and all videos are downsampled to 16 frames, there are fewer model parameters allowing the use of higher resolution video input.  Frames are resized to $640\times360$ pixels and $180\times180$ center cropping is used.\\
\textbf{Evaluation metrics:} AQA is assessed using Spearman's rank correlation, dive classification uses accuracy, and commentary uses captioning metrics of Bleu, Meteor, Rouge, and CIDEr.  
\subsection{Single-task vs. Multi-task approach}
\label{sec:exp_tasks}
\begin{table}[]
\small
\centering
\begin{tabular}{l|c|c}
\toprule    \multicolumn{1}{l|}{\textbf{Tasks}}                  & \textbf{\begin{tabular}[c]{c@{}}\textsc{C3D-AVG}\end{tabular}} & \textbf{\begin{tabular}[c]{c@{}}\textsc{MSCADC}\end{tabular}} \\ \midrule 
{AQA}          & 89.60                                                           & 84.72                                                          \\ \midrule 
{+ Cls}        & 89.62                                                           & 85.76                                                          \\
{+ Caps}       & 88.78                                                           & 85.47                                                          \\
{+ Cls + Caps} & \textbf{90.44}                                                  & \textbf{86.12} \\ \bottomrule                                             
\end{tabular}
\caption{\textbf{\textsc{STL} vs. \textsc{MTL} across different architectures}. Cls - classifiction, Caps - captioning. First row shows STL results, while the remaining rows show MTL results.
}
\label{tab:stl_vs_mtl}
\end{table}
We carry out an experiment to compare the performance of STL against that of MTL. We have a total of 3 tasks: AQA, detailed action recognition, and commentary generation. This experiment first considered the STL approach to AQA task and then measured the effect of including auxiliary tasks.  The evaluation is summarized in Table \ref{tab:stl_vs_mtl}. We observe that MTL approaches perform better than STL approach for both the networks, which shows that our MTL approach is not limited to a network but is generalizable across networks. Other thing to note here is that MTL performance improves as we incorporate more tasks. 
Comparing both the architectures, we find that our C3D-AVG outperforms our MSCADC for both STL and MTL, while MSCADC has the advantage of being fast and lower memory requirement than C3D-AVG. For qualitative results, refer to Table \ref{fig:qual_res} and supplementary material.  

\begin{table}[]
\small
\centering
\begin{tabular}{l|c}
\toprule
\multicolumn{1}{l|}{\textbf{Method}}    & \textbf{Sp. Corr.}    \\ \midrule
Pose+DCT \cite{pirsia}                       & 26.82        \\ 
C3D-SVR \cite{parmar17}                        & 77.16        \\ 
C3D-LSTM \cite{parmar17}                       & 84.89        \\ 
Ours MSCADC-STL                 & 84.72        \\ 
Ours C3D-AVG-STL                & \textbf{89.60}        \\ \midrule
Ours MSCADC-MTL                & 86.12        \\ 
Ours C3D-AVG-MTL               & \textbf{90.44}        \\ 
\multicolumn{2}{c}{} \\ 
\multicolumn{2}{c}{\textit{Segment-specific methods (train/test on UNLV Dive \cite{parmar17})}} \\ \midrule
S3D (best performing in \cite{s3d})       & 86.00        \\ 
Li \etal \cite{yongjun}                    & 80.09        \\ 
Ours MSCADC-STL					& 79.79		   \\
Ours C3D-AVG-STL                & 83.83        \\ \midrule
Ours MSCADC-MTL					& 80.60		   \\
Ours C3D-AVG-MTL               & \textbf{88.08}        \\ \bottomrule
\end{tabular}%
\caption{\textbf{Performance comparison with the existing AQA approaches}.}
\label{tab:sota}
\end{table}
\begin{table}[]
\small
\centering
\begin{tabular}{l|c|c|c}
\cmidrule[\heavyrulewidth]{2-4}
                        & \multirow{2}{*}{\begin{tabular}[c]{@{}c@{}}Nibali\\ \etal \cite{nibali}\end{tabular}} & \multicolumn{2}{c}{Ours-MTL}         \\ 
                        &                                                                                 & MSCADC & C3D-AVG \\ \midrule
\textbf{Position}       & 74.79                                                                           & 78.47               & \textbf{96.32}                \\
\textbf{Amstand}        & 98.30                                                                          & 97.45               & \textbf{99.72}                \\
\textbf{Rotation type}  & 78.75                                                                           & 84.70               & \textbf{97.45}                \\
\textbf{\# Somersaults} & 77.34                                                                           & 76.20               & \textbf{96.88}                \\
\textbf{\# Twists}      & 79.89                                                                           & 82.72               & \textbf{93.20}                \\ \bottomrule
\end{tabular}
 
\vspace{10px}
\begin{footnotesize}
\begin{tabular}{l|ccccccc}
\toprule
\textbf{Model}   & \textbf{B1} & \textbf{B2} & \textbf{B3} & \textbf{B4} & \textbf{M} & \textbf{R} & \textbf{C} \\ \midrule
C3D-AVG & 0.26         & 0.10         & 0.04         & 0.02         & 0.11       & 0.14       & 0.06       \\ 
MSCADC  & 0.25         & 0.09         & 0.03         & 0.01         & 0.11       & 0.13       & 0.05       \\ \bottomrule
\end{tabular}
\end{footnotesize}

\caption{\textbf{Performance on auxiliary tasks}.
}
\label{res:dive_cls}
\end{table}
\begin{table*}[t]
\footnotesize
\centering
\resizebox{\textwidth}{!}{%
\begin{tabularx}{\linewidth}{lm{14cm}}
\toprule 
\multirow{5}{*}{\animategraphics[loop,autoplay,poster=17,height=0.1\textwidth]{3}{clip2/22/}{0}{29}} & \textbf{\textsc{GT} labels:} 89.08; {[}Tuck, No, Backwards, 3.5, 0{]}\\  &\textbf{\textsc{C3D-AVG} labels:} 80.41; {[}Tuck, No, Backwards, 3.5, 0{]} \\
&\textbf{\textsc{MSCADC} labels:} 85.09; {[}Pike, No, Backwards, 2.5, 0{]} \\  
				   & \textbf{\textsc{GT} Caps:} Games but that shoulder issue it's not even paddling with that arm I hope he hasn't done something serious here and I think he has he looks even he is struggling he is in trouble and this is most concerning his hope he hasn't suffered a little dislocation there he's managed to get himself up onto the pool edge that looks like a dislocation to me what a way to end that's not the case but he's in trouble \\
                   & \textbf{\textsc{C3D-AVG} Caps:} that's good she will certainly keep that with a dive that we can certainly do it in that field very impressive it is very good but it was a good dive here playing a little bit low water which is a strong start one look at that closes vertical position it's gonna get seven and a half's I think super slow-motion just throwing up a little bit too much splash but she'll get over 60s mid sixties probably and that will be good enough to keep her middle of the pack which is where she needs to be to \\
                   & \textbf{\textsc{MSCADC} Caps:} well that's a really good dive from her Cheung just last night for that the dive and over she goes die are faltering and a good takeoff if she gets her feet are so she drills she's feet up and then she goes into the water with her chin on the end of the diving board on the end of the diving board squeezing into the Tuck shape as tight as you can see here we are here's the one and a half twist into the pike and that makes it down there have a lot \\ 
                  \midrule

\multirow{5}{*}{\animategraphics[loop,autoplay,poster=17,height=0.1\textwidth]{3}{clip2/38/}{0}{29}} & \textbf{\textsc{GT} labels:} 63.07; {[}Free, Yes, Backwards, 2, 2.5{]}\\  &\textbf{\textsc{C3D-AVG} labels:} 65.79; {[}Free, Yes, Backwards, 2, 1.5{]} \\ 
&\textbf{\textsc{MSCADC} labels:} 63.23; {[}Free, Yes, Backwards, 2, 1.5{]} \\
				   & \textbf{\textsc{GT} Caps:} it's beautiful our balance he is cranked the start a little bit too much he giving too much beans and could not stop that rotation it's a tricky dive to do it many times it's just a real pain because you've got so many twists to do to get in with a rotation it's very easy to over rotate that's exactly what Mitchum's done he is lagging behind now this could prove to be critical dropped another 12 \\
                   & \textbf{\textsc{C3D-AVG} Caps:} well it's okay and the entry into the water not quite 100\% vertical but he's just a little bit overcooked on the end obviously there's a few of the divers have you use themselves so if you're getting your hands out there we are rocking and rolling a little bit of a splash with technically a little bit of splash that's not the splash means that the judges will penalize him or only got to 17 from the two and a half somersaults before he goes into the water now that was a \\ 
                   & \textbf{\textsc{MSCADC} Caps:} well that's a really good dive yeah it's a good dive from enough so she needed to be faltering she knows to go in her hair it's a difficult dive to finish in the business end of the field and away in the pike position this is the reverse two and a half there she didn't get around but instead of she goes wrong with her hands on I think she has done very well I'm not surprised as Paul Espinoza she gets around the 310 mark maybe yeah she will do well do \\ 
                 \midrule

\multirow{5}{*}{\animategraphics[loop,autoplay,poster=17,height=0.1\textwidth]{3}{clip2/68/}{0}{29}} & \textbf{\textsc{GT} labels:} 84.15; {[}Tuck, No, Backwards, 3.5, 0{]} \\& \textbf{\textsc{C3D-AVG} labels:} 81.94; {[}Tuck, No, Backwards, 3.5, 0{]} \\
&\textbf{\textsc{MSCADC} labels:} 87.04; {[}Tuck, No, Backwards, 3.5, 0{]} \\  
				   & \textbf{\textsc{GT} Caps:} states very good seventy nine point two for the dive last night brings a smile to the US coaching staff great height off the platform so he's already threw his first somersault before he passes it on the way down give yourself time to get the entry angle just right which he did started off as a swimmer and said that to switch to diving because he looked more fun having a bit more fun \\
                   & \textbf{\textsc{C3D-AVG} Caps:} excellent excellent dive if you might he's got a lot of divers here with their hands together for him a lot of them here and take a little bit of an angle on the entry that does good through that would not quite a way over a vertical look at that perfect angle so much better judges will like that that angle so not too many \\
                   & \textbf{\textsc{MSCADC} Caps:} well that's a really good dive yeah it's a good dive from marginally short she goes in but it's not going to get more than a couple of those three point seven as you can see here we are here's to show diving consistently and here but she didn't falter but she had enough mediocre dives on the dive you need to be doing the back of the head there you know you need to flick it and then just about that toes on the head on this dive is there when you know the competition's are coming from \\ 
                 \midrule

\multirow{5}{*}{\animategraphics[loop,autoplay,poster=17,height=0.1\textwidth]{3}{clip2/71/}{0}{29}} & \textbf{\textsc{GT} labels:} 47.77; {[}Pike, No, Forwards, 2.5, 1{]} \\& \textbf{\textsc{C3D-AVG} labels:} 53.04; {[}Pike, No, Forwards, 2.5, 1{]} \\
&\textbf{\textsc{MSCADC }labels:} 67.98; {[}Pike, No, Backwards, 2.5, 0{]} \\  
				   & \textbf{\textsc{GT} Caps:} final but she has benefited from the experience and again another diver going a long long way down the pool and away from the platform yeah just it's the technique if you haven't got the right timing on the jump and momentum of the run down the board just drags you forward high score so far is 46 there we \\
                   & \textbf{\textsc{C3D-AVG} Caps:} nice nice entry because the execution was fine and then just suggesting she went surfing over the end of the diving board anyway she's a safe distance from the diving board so that's a good dive in the prelims you can see the splash moving away from the diving board six and a half's sevens at best moving further away from the podium dive after dive star with a 58 and it with a 64 this \\ 
                   & \textbf{\textsc{MSCADC} Caps:} well that's a busy dive finish from the end but she certainly was better than the three point three she didn't really attack she was her fourth round dive she knows too far away from the diving board squeezing into the Tuck shape just bending at a little bit too early they're short of vertical an injury little bit too much splash but she'll be lucky by the toes on the water the divers would have been that towel on the end of the divers do that the divers would have touched the scoreboard but I think she's gonna \\ 
                \bottomrule
\end{tabularx}
}
\caption{\textbf{Qualitative results}. Labels are ordered as follows: AQA score; [Position, Armstand?, Rotation type, \#SS, \#TW].}
\label{fig:qual_res}
\end{table*}
Next, we compare our models with the existing methods in Table \ref{tab:sota}. We obtain the results for all of the existing methods on our dataset. C3D-SVR was the best performing method in \cite{parmar17} but it does not seem to benefit from the increased number of training samples. In \cite{parmar17}, C3D-LSTM was reported to be performing worse than C3D-SVR due to insufficient amount of training data and does outperform C3D-SVR with the expanded training data. Our MSCADC-STL works better than most of the existing methods, whereas our C3D-AVG-STL is better performing than all the existing methods. Furthermore, C3D-AVG-MTL with 90.44\% correlation achieves new state-of-the-art results.

Method proposed by Xiang \etal \cite{s3d} requires manual annotation to mark end points of all the segments which is not available in the new Diving-MTL data.  Xiang \etal \cite{s3d} used the UNLV-Dive dataset \cite{parmar17} so for a fair comparison with \cite{s3d} we train and test our models on UNLV-Dive \cite{parmar17}. The results are enumerated in Table \ref{tab:sota}. Our C3D-AVG-STL does not perform as well S3D \cite{s3d}. However, our C3D-AVG-MTL outperforms the S3D model. An important thing to note here is that UNLV-Dive dataset is quite a bit smaller than our newly introduced MTL-AQA dataset which should limit MTL performance. However, as pointed out in Section \ref{sec:approach}, MTL provides better generalization than STL, which allows C3D-AVG-MTL to learn effectively from fewer training samples.

Performance on the auxiliary tasks is presented in Table \ref{res:dive_cls}.  To the best of our knowledge there is only one work (by Nibali \etal \cite{nibali}) on detailed dive classification. Our C3D-AVG-MTL performed best on the classification task as well. We also give captioning metrics for the two networks though there is no baseline for comparison in literature.  
\begin{table}[]
\small
\centering
\begin{tabular}{l|cccc}
\toprule
\textbf{\# samples} & \textbf{1059}           & \textbf{450}            & \textbf{280}            & \textbf{140}            \\
\midrule
\textsc{STL}        & 89.60          & 77.27          & 69.63          & 64.17          \\
\textsc{MTL}        & \textbf{90.44} & \textbf{83.52} & \textbf{72.09} & \textbf{68.16} \\
\bottomrule
\end{tabular}
\caption{\textbf{STL vs. MTL generalization}. Training using increasingly reduced no. of training samples.}
\label{res:no_train_samples}
\end{table}
\paragraph{Generalization provided by MTL: }
To ascertain that MTL is providing more generalization, we train our C3D-AVG-STL and C3D-AVG-MTL models using fewer number of datapoints. Train set size and the corresponding STL/MTL performances are detailed in Table \ref{res:no_train_samples}. We see that MTL consistently outperforms STL, and also the gap seems to widen with fewer training samples.
\subsection{AQA-orientedness of the learned \\ representations}
\label{sec:exp_layers}
\begin{table}[]
\small
\centering
\setlength\tabcolsep{4pt}%
\begin{tabular}{l|ccccc}
\cmidrule[\heavyrulewidth]{2-6}
      & \tt{\textbf{c1}}    & \tt{\textbf{c2}}    & \tt{\textbf{c3}}    & \tt{\textbf{c4}}    & \tt{\textbf{c5}}    \\ \midrule
Baseline-1          & 71.01          & 71.39          & 73.13          & 76.34          & 73.69          \\
Baseline-2			& 72.43		& 70.15		& 70.35 	& 57.20 	& 37.63 \\
\textsc{C3D-AVG-MTL} & \textbf{74.26}          & \textbf{77.95}          & \textbf{82.78} & \textbf{86.18} & \textbf{85.75} \\ \bottomrule
\end{tabular}
\caption{\textbf{Performance of fitting linear regressors on the activations of all the convolutional layers}.}
\label{res:lin_reg}
\end{table}
\begin{table}[]
\small
\centering
\setlength\tabcolsep{3pt}%
\begin{tabular}{l|ccccc}
\cmidrule[\heavyrulewidth]{2-6}
         & \tt{\textbf{c1}}          & \tt{\textbf{c2}}          & \tt{\textbf{c3}}          & \tt{\textbf{c4}}          & \tt{\textbf{c5}}          \\ \midrule
\multicolumn{6}{c}{\textit{Train/Test events overlapping}}                                                                              \\ 
Baseline-1            & \textbf{41.10}       & 32.06                & 36.53                & 46.86                & \textbf{44.78}       \\ 
Baseline-2	& 37.76		& 42.02 	& 37.98		& 44.28 	& 38.56 \\
C3D-AVG-MTL       & 38.32                & \textbf{42.68}                & \textbf{45.53}       & \textbf{49.18}       & 38.47            \\ \midrule
\multicolumn{6}{c}{\begin{tabular}[c]{@{}c@{}}\textit{Train/Test events non-overlapping}\\ \textit{(requires more generalization)}\end{tabular}} \\ 
Baseline-1           & -02.68               & 00.75                & -03.91               & -02.22               & 03.17                \\ 
Baseline-2 		& -07.52 	& -02.44 	& 05.07		& 24.09 	& \textbf{25.80} \\ 
C3D-AVG-MTL       & -07.75               & -02.77               & \textbf{23.51}       & \textbf{29.56}       & -03.25               \\ \bottomrule
\end{tabular}
\caption{\textbf{Performance of fitting linear regressors on the activations of all the convolutional layers for a novel action class, Gymnastic vault}. Top rows: Within-dataset evaluation, bottom rows: Out-of-dataset evaluation.}
\label{res:zeroshot}
\end{table}
We trained our networks end-to-end to learn AQA-specific feature representation rather than relying on pretrained action-recognition oriented features (as done in \cite{parmar17}). However, we question if there is a utility in learning AQA-specific feature representation or are action-recognition oriented features equally good? To answer this, we follow an evaluation scheme similar to Zhang \etal \cite{zhang16}, where we train linear regressors on top of all the convolutional layers, and compare the performance obtained for AQA and action-recognition models. In particular, we consider two action-recognition baselines: C3D model trained on UCF-101 dataset \cite{ucf101} (Baseline-1), and our model trained on our MTL-AQA dataset, but for factorized action recognition task (Baseline-2).

In the primary evaluation, we compare the representations for measuring the quality of diving action. Comparison is detailed in Table \ref{res:lin_reg}. In comparison to both the baselines, we find that our C3D-AVG-MTL learns better representations at all the intermediate layers. 

Further we compare the representations for measuring the quality of an unseen action class -- Gymnastic vault \cite{parmar17}. This helps in estimating the generalizability of the representations. We hypothesize that if our AQA network has learned better representations that actually capture the concept of \emph{quality} in an action, then it should be able to measure the quality of an unseen action better than action-recognition specific networks. We carry out 2 different evaluations: 1) \textbf{Within-dataset evaluation} and 2) \textbf{Out-of-dataset evaluation}. In Within-dataset evaluation we randomly divide the samples into train set and test set, whereas in Out-of-dataset evaluation, train and test samples are drawn from different athletic competitions. Out-of-dataset evaluation is more challenging and requires feature representations to be more generalizable and not suffer from dataset-bias. Like the previous experiment, to compare learned representations, we train linear regressors on top of all the convolutional layers. Train and test sets consist of 125 and 56 samples respectively. Results from both evaluations are presented in Table \ref{res:zeroshot}. 

In the Within-dataset evaluation, the representations learned by all the models seem to be working well, although C3D-AVG-MTL performs best. The difference in performance becomes clearer in the Out-of-dataset evaluation. As expected, Out-of-dataset evaluation is more challenging and performances of all the models drop. However, the performances of Baseline-2 and our model drop more gracefully.
\section{Discussion}
\label{sec:discussion}
We introduced a multitask learning approach to AQA and showed that MTL performs better than STL because of better generalization which is especially important in AQA and skill assessment since datasets are small. We showed that the representations learned by our MTL models are better able to capture the inherent concept of quality of actions. Our approach is scalable since the supervision required for the auxiliary tasks is readily available from the existing video footage with minimal extra effort compared to just AQA labeling. In addition, state-of-the-art performance was achieved without any finetuning of hyperparameters.  Our best performing and recommended model, C3D-AVG-MTL, achieved 90.44\% correlation with judged scores which still leaves a small gap to achieve human-experts-level performance (96\% \cite{pirsia}). 
\vspace{-0.1cm}
\paragraph{Extension to other actions and skills assessment:} Although this paper is geared specifically toward multitask diving AQA, the approach is general in nature.  No design decisions were biased towards or specific to the diving tasks.  Experiments even showed that the models trained on diving do work reasonably well for another action, gymnastic vault.  This encouraging result hints at the direct application of our MTL approach on other actions and everyday skills assessment. Commentary and action class details are available almost all the of time in the sport footages.  For non-sport skills assessment, such as surgery, needle passing, drawing, or painting, experts could be used to generate comments and definition of sub-actions for classification.  Note that existing datasets can simply be augmented to include additional labels, instead of building new datasets from scratch. Also, our MTL approach is complementary to the existing AQA and skills assessment approaches. 
\paragraph{Acknowledgements:} We would like to thank Andy Squadra, Mark Wilbourne, Josh Rana for helping us with the dataset collection.
{\small
\bibliographystyle{ieee_fullname}
\bibliography{egbib}
}
\end{document}